\documentclass[conference]{IEEEtran}
\IEEEoverridecommandlockouts
\usepackage{cite}
\usepackage{amsmath,amssymb,amsfonts}
\usepackage{algorithm}
\usepackage{algpseudocode}
\usepackage{graphicx}
\usepackage{physics}
\usepackage{textcomp}
\usepackage{hyperref}
\usepackage{xcolor}
\usepackage{comment}
\usepackage{caption}
\usepackage{subcaption}
\def\BibTeX{{\rm B\kern-.05em{\sc i\kern-.025em b}\kern-.08em
    T\kern-.1667em\lower.7ex\hbox{E}\kern-.125emX}}

\hypersetup{
    colorlinks=true,
    citecolor=black,
    linkcolor=black,
    filecolor=magenta,      
    urlcolor=blue,
}

\DeclareMathOperator*{\minimize}{minimize}

\newcommand{\boldx}[0]{\mathbf{x}}

\newcommand{\pos}[2]{x_{#1}^{(#2)}}
\newcommand{\vel}[2]{v_{#1}^{(#2)}}
\newcommand{\mom}[2]{M_{#1}^{(#2)}}
\newcommand{\pbest}[2]{pbest_{#1}^{(#2)}}
\newcommand{\gbest}[1]{gbest^{(#1)}}

\newcommand{\chisup}[1]{\chi^{(#1)}}
\newcommand{\Lim}[1]{\raisebox{0.5ex}{\scalebox{0.8}{$\displaystyle \lim_{#1}\;$}}}
\newcommand{\prob}[1]{p_{#1}(#1)}

\newcommand{\sub}[1]{ $_{#1}$}
\newcommand{\subzero}[0]{\sub{0.00}}
    
\begin{document}

\title{Fairly Constricted Multi-Objective Particle Swarm Optimization
\thanks{Project is funded by DST}
}

\author{\IEEEauthorblockN{1\textsuperscript{st} Anwesh Bhattacharya}
\IEEEauthorblockA{\textit{Dept. of Physics \& CSIS} \\
\textit{BITS-Pilani, Pilani}\\
Pilani, India \\
f2016590@pilani.bits-pilani.ac.in}
\and
\IEEEauthorblockN{2\textsuperscript{nd} Snehanshu Saha}
\IEEEauthorblockA{\textit{Dept. of CSIS \& APPCAIR} \\
\textit{BITS-Pilani, Goa}\\
Goa, India \\
snehanshus@goa.bits-pilani.ac.in}
\and
\IEEEauthorblockN{2\textsuperscript{nd} Nithin Nagaraj}
\IEEEauthorblockA{\textit{Consciousness Studies Programme} \\
\textit{NIAS}\\
Bangalore, India \\
nithin@nias.res.in}
}

\maketitle

\begin{abstract}
It has been well documented that the use of exponentially-averaged momentum (EM) in particle swarm optimization (PSO) is advantageous over the vanilla PSO algorithm. In the single-objective setting, it leads to faster convergence and avoidance of local minima. Naturally, one would expect that the same advantages of EM carry over to the multi-objective setting. Hence, we extend the state of the art Multi-objective optimization (MOO) solver, SMPSO, by incorporating EM in it. As a consequence, we develop the mathematical formalism of \emph{constriction fairness} which is at the core of extended SMPSO algorithm. The proposed solver matches the performance of SMPSO across the ZDT, DTLZ and WFG problem suites and even outperforms it in certain instances.
\end{abstract}


\section{Introduction}
\label{sec:intro}

\subsection{Vanilla PSO}

Particle Swarm Optimization (PSO) was first proposed by Kennedy and Eberhart \cite{kennedy.eberhart.og, swarm.intelligence} in 1995 as an evolutionary single-objective optimization algorithm. $N$ particles are initialised at random positions/velocities in the search space, and the $i^{\text{th}}$ particle updates its trajectory according to
\begin{align}
    \vel{i}{t+1} &= w\vel{i}{t} + c_1r_1(\pbest{i}{t} - \pos{i}{t}) \nonumber \\
    &+ c_2r_2(\gbest{t} - \pos{i}{t}) \label{eq:updatev} \\
    \pos{i}{t+1} &= \pos{i}{t} + \vel{i}{t+1} \label{eq:updatex}
\end{align}
$r_1$ and $r_2$ are random numbers drawn from the uniform distribution $U(0,1)$. $\pbest{i}{t}$ is the best position (in terms of minimizing the objective) that particle $i$ has visited upto time $t$. $\gbest{t}$ is the best position among all particles that has been achieved. After sufficient iterations, all particles assume positions $x_i$ near $gbest$ with the particle velocities $v_i\approx0$. In this state, we say that the swarm has converged. 

\subsection{Exponentially-Averaged Momentum}

\cite{adaswarm} proposes the EMPSO algorithm to speed up the convergence and avoid local minima in single-objective problems. It is a vanilla PSO algorithm aided by exponentially-averaged momentum (EM). Their PSO update equations are as follows 
\begin{align}
    \mom{i}{t+1} &= \beta\mom{i}{t} + (1-\beta)\vel{i}{t} \label{eq:updatem.mom} \\
    \vel{i}{t+1} &= \mom{i}{t+1} + c_1r_1(\pbest{i}{t} - \pos{i}{t}) \nonumber \\
    &+ c_2r_2(\gbest{t} - \pos{i}{t}) \label{eq:updatev.mom}
\end{align}
Eq (\ref{eq:updatem.mom}) computes the exponentially-averaged velocity of the $i^{\text{th}}$ particle upto timestep $t$. The position update equation for EMPSO remains the same as eq (\ref{eq:updatex}). The momentum factor must obey $0<\beta<1$.\footnote{Note that $\beta=0$ degenerates to vanilla PSO} By recursively expanding eq (\ref{eq:updatem.mom}), a particle's momentum is an exponentially weighted sum of all its previous velocities
\begin{align}
    \mom{i}{t+1} &= (1-\beta)\vel{i}{t}  + \beta (1 - \beta)\vel{i}{t-1} + \ldots \nonumber \\
    &+ \beta^{t-2}(1 - \beta)\vel{i}{2} + \beta^{t-1} (1-\beta)\vel{i}{1}
\end{align}
In certain single-objective problems, \cite{adaswarm} report a $50\%$ reduction in the iterations taken to convergence for EMPSO relative to vanilla PSO. Due to its superior performance over the vanilla algorithm in the single-objective setting, we hypothesize that similar benefits of EM would be seen in multi-objective problems. 

\subsection{Multi-Objective Optimization}

The central setting of multi-objective optimization (MOO) is the following problem
$$\minimize_{\boldx\in \mathbb{R}^n} \mathbf{f}(\mathbf{x}) = [f_1(x), f_2(x), \ldots, f_k(x)]$$
\emph{i.e.,} given an input space $\mathbb{R}^n$, we want to optimize $k$ functions $f_1, f_2, \ldots, f_k$ in the objective space. In practice, MOO solvers find an \emph{Pareto front} which represents a non-dominated set of decision variables $\boldx_i\in\mathbb{R}^n$. Simplistically, it is a set of solutions where each member of the set is as good a solution as any other. A comprehensive introduction to MOO can be found in \cite{intro.moo}. SMPSO\cite{smpso} is the state-of-the-art MOO solver that is based on vanilla PSO. It uses a constricted vanilla PSO whose update equation is
\begin{align}
    &\vel{i}{t+1} = \chi[w\vel{i}{t} + \nonumber \\
    &c_1r_1(\pbest{i}{t} - \pos{i}{t}) + \nonumber \\
    &c_2r_2(\gbest{t} - \pos{i}{t})] \label{eq:updatev.constrict} 
\end{align}
where $\chi$ is the constriction factor \cite{explosion} defined as follows
\begin{gather}
\chi =
\left\{
	\begin{array}{ll}
		\frac{2}{2-\phi-\sqrt{\phi^2-4\phi}}  & \mbox{ } \phi > 4 \\
		1 & \mbox{ } \phi \leq 4
	\end{array}
\right. \label{eq:constriction.pso}
\end{gather}
with $\phi=c_1+c_2$. Hence $\chi$ is a function of $c_1,c_2$. Since the constriction factor is with respect to vanilla PSO, we denote it as $\chi\equiv\chisup{v}(\phi)$
\footnote{Note that constriction factor is negative for $\phi>4$}. The position update equation for constricted vanilla PSO remains the same as eq (\ref{eq:updatex}). We describe SMPSO in algorithm \ref{alg:smpso}. 
\begin{algorithm}. 
\caption{SMPSO Pseudocode}\label{alg:smpso}
\begin{algorithmic}[1]
\State $\mathit{initializeSwarm}$() \label{alg:smpso-line:initSwarm}
\State $leaders = \mathit{initializeArchive}()$ \label{alg.smpso-line:initArchive}
\State $gen \gets 0$
\While{$gen < maxGen$}
    \State $\mathit{computeSpeed}$() \label{alg:smpso-line:computeSpeed}
    \State $\mathit{updatePosition}$() \label{alg:smpso-line:updatePos}
    \State $\mathit{mutation}$() \label{alg:smpso-line:mutation}
    \State $\mathit{evaluation}$() \label{alg:smpso-line:eval}
    \State $\mathit{updateArchive}$(leaders)
    \State $\mathit{updateParticlesMemory}$() \label{alg:smpso-line:updatemem}
    \State $gen \gets gen + 1$
\EndWhile
\State \textbf{Return} $leaders$
\end{algorithmic}
\end{algorithm}

Line~\ref{alg:smpso-line:initSwarm} initializes the particle's positions in the input space along with random velocities. As per \cite{intro.moo}, the external archive for storing leaders is initialized in line~\ref{alg.smpso-line:initArchive}. Line~\ref{alg:smpso-line:computeSpeed} updates the swarm obeying constricted Vanilla PSO equations (\ref{eq:updatev.constrict}, \ref{eq:constriction.pso}). Line~\ref{alg:smpso-line:updatePos} follows the regular position update equation as eq (\ref{eq:updatex}). Line~\ref{alg:smpso-line:mutation} performs a turbulence mutation which introduces a diversity of solutions in the swarm, so that they don't converge to a single point. Finally, the particles are evaluated and the external archive is updated in lines \ref{alg:smpso-line:eval}-\ref{alg:smpso-line:updatemem}. In particular, we focus on line 5 of algorithm \ref{alg:smpso} and expand it in algorithm \ref{alg:smpso.computespeed}.
\begin{algorithm}. 
\caption{SMPSO $\mathit{computeSpeed}$()}\label{alg:smpso.computespeed}
\begin{algorithmic}[1]
\For{$i \gets 1$ to $swarmSize$}
    \State $r_1 \gets \mathit{Uniform}(0, 1)$ \label{alg:smpso.computespeed-line:r1}
    \State $r_2 \gets \mathit{Uniform}(0, 1)$ \label{alg:smpso.computespeed-line:r2}
    \State $c_1 \gets \mathit{Uniform}(1.5, 2.5)$ \label{alg:smpso.computespeed-line:c1}
    \State $c_2 \gets \mathit{Uniform}(1.5, 2.5)$ \label{alg:smpso.computespeed-line:c2}
    \State $\phi \gets c_1+c_2$ \label{alg:smpso.computespeed-line:phi}
    \State $\chi \gets
    \mathit{ConstrictionFactor}(\phi)$ \label{alg:smpso.computespeed-line:chi}
    \State $v[i] \gets wv[i] + c_1r_1(pbest[i]-x[i]) + c_2r_2(gbest-x[i])$\label{alg:smpso.computespeed-line:vi}
    \State $v[i] \gets \chi v[i]$ \label{alg:smpso.computespeed-line:chi.vi}
    \State $v[i]\gets \mathit{VelocityConstriction}(v[i])$ \label{alg:smpso.computespeed-line:vi.constrict}
\EndFor
\end{algorithmic}
\end{algorithm}

Lines \ref{alg:smpso.computespeed-line:r1}-\ref{alg:smpso.computespeed-line:r2} draw $r_1,r_2$ from a uniform distribution $U(0,1)$ and lines \ref{alg:smpso.computespeed-line:c1}-\ref{alg:smpso.computespeed-line:c2} draw $c_1, c_2\sim U(1.5, 2.5)$. Line \ref{alg:smpso.computespeed-line:phi} computes $\phi$ and line \ref{alg:smpso.computespeed-line:chi} computes the constriction factor $\chisup{v}(\phi)$.  Lines \ref{alg:smpso.computespeed-line:vi}-\ref{alg:smpso.computespeed-line:chi.vi} update the particles velocity according to eq (\ref{eq:updatev.constrict}) where $x[i]$ and $v[i]$ are the position and velocity vectors respectively of the $i^{\text{th}}$ particle. Finally, line \ref{alg:smpso.computespeed-line:vi.constrict} performs a velocity constriction based on the boundary of the search space\footnote{Details of this step can be found in \cite{smpso}}

SMPSO claims that its superiority over other MOO solvers, such as OMOPSO\cite{omopso} and NSGA-II\cite{nsgaii}, is rooted in the randomized selection of $c_1, c_2$ along with the constriction factor $\chisup{v}(\phi)$ which maintains a diversity of solutions in the swarm.  

\section{Motivations}

\subsection{An EM-aided SMPSO Algorithm}

Apart from the external archive, leader selection and mutation, the performance of SMPSO is governed by the dynamics of the swarm which is solely dictated by the $\mathit{computeSpeed}()$ subroutine (Algorithm \ref{alg:smpso.computespeed}). Thus, the incorporation of EM in SMPSO must occur within the $\mathit{computeSpeed}()$ function (line \ref{alg:smpso-line:computeSpeed} of Algorithm \ref{alg:smpso}). As a first attempt, we formulate the desired $\mathit{computeSpeed()}$ in Algorithm \ref{alg:emsmpso.computespeed}. We name our EM-aided SMPSO algorithm as \emph{EM-SMPSO}.
\begin{algorithm}. 
\caption{EM-SMPSO $\mathit{computeSpeed}$()}\label{alg:emsmpso.computespeed}
\begin{algorithmic}[1]
\For{$i \gets 1$ to $swarmSize$}
    \State $r_1 \gets \mathit{Uniform}(0, 1)$
    \State $r_2 \gets \mathit{Uniform}(0, 1)$
    \State $c_1 \gets \mathit{Uniform}(1.5, 2.5)$
    \State $c_2 \gets \mathit{Uniform}(1.5, 2.5)$
    \State $\beta \gets \mathit{Uniform}(0, 1)$ \label{alg:emsmpso.computespeed-line:beta}
    \State $\phi \gets c_1+c_2$
    \State $\chi \gets \mathit{ConstrictionFactor}(\phi, \beta)$ \label{alg:emsmpso.computespeed-line:chi}
    \State $m[i] \gets \beta m[i] + (1-\beta)v[i]$ \label{alg:emsmpso.computespeed-line:update1}
    \State $v[i] \gets m[i] + c_1r_1(pbest[i]-x[i]) + c_2r_2(gbest-x[i])$
    \State $v[i] \gets \chi v[i]$ \label{alg:emsmpso.computespeed-line:update2}
    \State $v[i]\gets \mathit{VelocityConstriction}(v[i])$
\EndFor
\end{algorithmic}
\end{algorithm}

Akin to Algorithm \ref{alg:smpso.computespeed}, we draw $\beta\sim U(0,1)$ in line~\ref{alg:emsmpso.computespeed-line:beta}. Line~\ref{alg:emsmpso.computespeed-line:chi} computes the appropriate constriction factor for EM-SMPSO. Note that the function $\mathit{ConstrictionFactor}$ now takes two arguments $(\phi,\beta)$ instead of one. This is because EM directly affects the swarm dynamics and hence we need a different constriction factor $\chi\equiv\chisup{m}(\phi,\beta)$. Lines~\ref{alg:emsmpso.computespeed-line:update1}-\ref{alg:emsmpso.computespeed-line:update2} are the update equations of constricted EMPSO.

\subsection{Constriction Factor for EMPSO}

We follow the derivation of the constriction factor according to \cite{explosion}. Consider a \emph{determinisitc} version of Vanilla PSO equations with $[x, v]$ as a 2-D discrete-time map (equations obtained from \cite{explosion})
\begin{align}
    v(t+1) &= v(t) + \phi(g-x(t)) \label{eq:constrict.pso.vtime} \\
    x(t+1) &= x(t) + v(t+1) \label{eq:constict.pso.xtime}
\end{align}
This is deterministic because we have fixed $p_{best}=g_{best}=g$. Let $y(t) = g - x(t)$ and we choose $w=1$ for simplicity. If we introduce a momentum time-series $m(t)$, we get a 3-D discrete-time map in $[v, y, m]$ for EMPSO as follows
\begin{align}
    v(t+1) &= (1-\beta)v(t) + \phi y(t) + \beta m(t) \label{eq:constrict.empso.v} \\
    y(t+1) &= (\beta - 1)v(t) + (1-\phi)y(t) - \beta m(t) \label{eq:constrict.empso.y} \\
    m(t+1) &= (1-\beta)v(t) + \beta m(t) \label{eq:constrict.empso.m}
\end{align}
The evolution matrix of this system is
\begin{gather}
U = \begin{bmatrix}
1-\beta & \phi & \beta \\
\beta-1 & 1-\phi & -\beta \\
1-\beta & 0 & \beta
\end{bmatrix} \label{eq:evolution.matrix}
\end{gather}
The eigenvalues of the matrix $U$ provide important information about the swarm dynamics \cite{strogatz}. According to \cite{explosion}, we are interested in eigenvalues $|\lambda|>1$ for deriving a constriction factor
\begin{gather}
    \lambda_{\pm} = \frac{(2-\phi) \pm \sqrt{\phi^2-4(1-\beta)\phi}}{2} \label{eq:eigen.sols}
\end{gather}
\cite{explosion} mentions that constriction entails finding the scaling factor $\chi$ for the eigenvalues $\lambda$ such that setting $\lambda'=\chi\lambda$ gives $|\lambda'| \leq 1$. Hence it is sufficient to set
\begin{gather}
    \chi = \frac{1}{max(|\lambda_+|, |\lambda_-|)} \label{eq:chi.sufficient}
\end{gather}
where $\lambda_{\pm}$ is either of the roots defined in eq (\ref{eq:eigen.sols}). Based on the discriminant $\Delta=\phi^2-4(1-\beta)\phi$, we obtain two cases that give us real/complex roots for $\lambda$.

\noindent\textbf{Case 1} ($\Delta \leq 0$) ---
\begin{align}
    |\lambda| &= \sqrt{
    \frac{(2-\phi)^2 - \Delta}{4}
    } \nonumber \\
    &= \sqrt{1-\beta \phi} \label{eq:eigen.mod}
\end{align}
Since $\sqrt{1-\beta \phi}$ is an absolute modulus, we must have
\begin{gather}
    \phi \leq \frac{1}{\beta} \label{eq:inverse.beta}
\end{gather}
Moreover, $\Delta \leq 0 \implies \phi \leq 4(1-\beta)$. Also, eq (\ref{eq:inverse.beta}) must be satisfied simultaneously. To check whether this is true, construct a function in the range $\beta \in (0,1)$
\begin{gather}
    f(\beta) = \frac{1}{\beta} - 4(1-\beta) \label{eq:respect.def}
\end{gather}
and its derivative
\begin{gather}
    f'(\beta) = 4 - \frac{1}{\beta^2} \label{eq:respect.der}
\end{gather}
The critical point is $\beta=\frac{1}{2}$ which is also the global minimum in $\mathbb{R}^+$ due to $f''(\beta) = \frac{2}{\beta^3} > 0$. Moreover, from $f(\frac{1}{2})=0$, we have $f(\beta)\geq0$ in its domain and
\begin{align}
    & \frac{1}{\beta} \geq 4(1-\beta)  \nonumber \\
    \implies & \phi \leq 4(1-\beta) \leq \frac{1}{\beta} \nonumber
\end{align}
Eq (\ref{eq:inverse.beta}) is thus satisfied and $|\lambda|=\sqrt{1-\beta\phi}\leq1$, hence we can set $\chi=1$.

\noindent \textbf{Case 2} ($\Delta > 0$) ---
Define $\lambda_m=max(|\lambda_{\pm}|)$. It can be shown that it simplifies to
\begin{align}
    \lambda_m = \frac{|\phi-2|+\sqrt{\phi^2-4(1-\beta)\phi}}{2} \label{eq:lambda.m}
\end{align}
Since we are interested in $\lambda_m>1$, we can do
\begin{align}
    \lambda_m^2 &> 1 & \nonumber \\
    \implies 4\beta\phi + 4|\phi-2| &> 8 \label{eq:phi2.ineq}
\end{align}

\noindent \textbf{Subcase 2.1} ($\phi > 2$) --- Eq (\ref{eq:phi2.ineq}) simplifies to
\begin{align}
    \phi>4(1+\beta)^{-1} \label{eq:constrict.ineq2}
\end{align}
For simplicity, we denote $\omega=4(1+\beta)^{-1}$. $\beta$ exists in the interval $0 < \beta < 1$. This implies that $2 < \omega < 4$. Thus, there exists an interval of $\beta, \phi$ where eq (\ref{eq:constrict.ineq2}) is satisfied for a suitable interval of $\phi$. If we choose $\phi \in [2, 4]$, this condition can be met.

\noindent \textbf{Subcase 2.2} ($\phi \leq 2$) --- It can be shown that eq (\ref{eq:phi2.ineq}) leads to $\beta>1$ which is a contradiction. Hence, this subcase can be ignored.

For the sake of implementation, we adopt a negative constriction co-efficient as has been used in SMPSO for our formulation. Hence, we combine eq (\ref{eq:chi.sufficient}, \ref{eq:lambda.m}) and eq (\ref{eq:constrict.ineq2})
\begin{gather}
\chisup{m}(\phi,\beta) =
\left\{
	\begin{array}{ll}
		\frac{2}{2-\phi-\sqrt{\phi^2-4(1-\beta)\phi}}  & \mbox{ } \phi > 4(1+\beta)^{-1} \\
		1 & \mbox{ } \text{otherwise}
	\end{array}
\right. \label{eq:constriction.empso}
\end{gather}
From a theoretical standpoint, adopting a positive/negative constriction co-efficient are equivalent because only the modulus $|\lambda|$ is significant \cite{strogatz}. Moreover, note that $\beta=0$ implies that the effect of momentum is absent and it can be easily confirmed that $\chisup{m}(\phi, 0)=\chisup{v}(\phi)$. Thus, our derivation is consistent with that of vanilla PSO.

\subsection{Preliminary Results}
\label{sec:prelim}

\begin{figure}[htbp]
     \centering
     \begin{subfigure}[b]{0.24\textwidth}
         \centering
         \includegraphics[width=\textwidth]{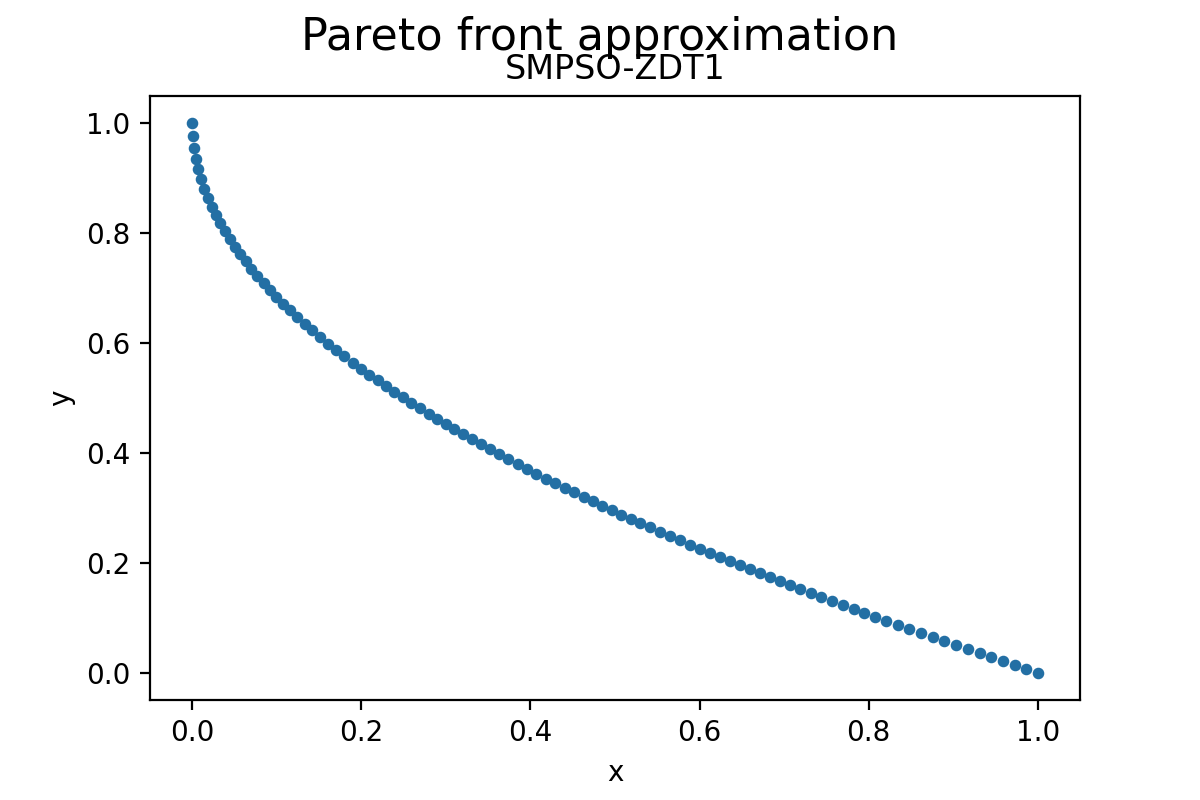}
         \caption{SMPSO}
         \label{fig:smpso.zdt1}
     \end{subfigure}
     \begin{subfigure}[b]{0.24\textwidth}
         \centering
         \includegraphics[width=\textwidth]{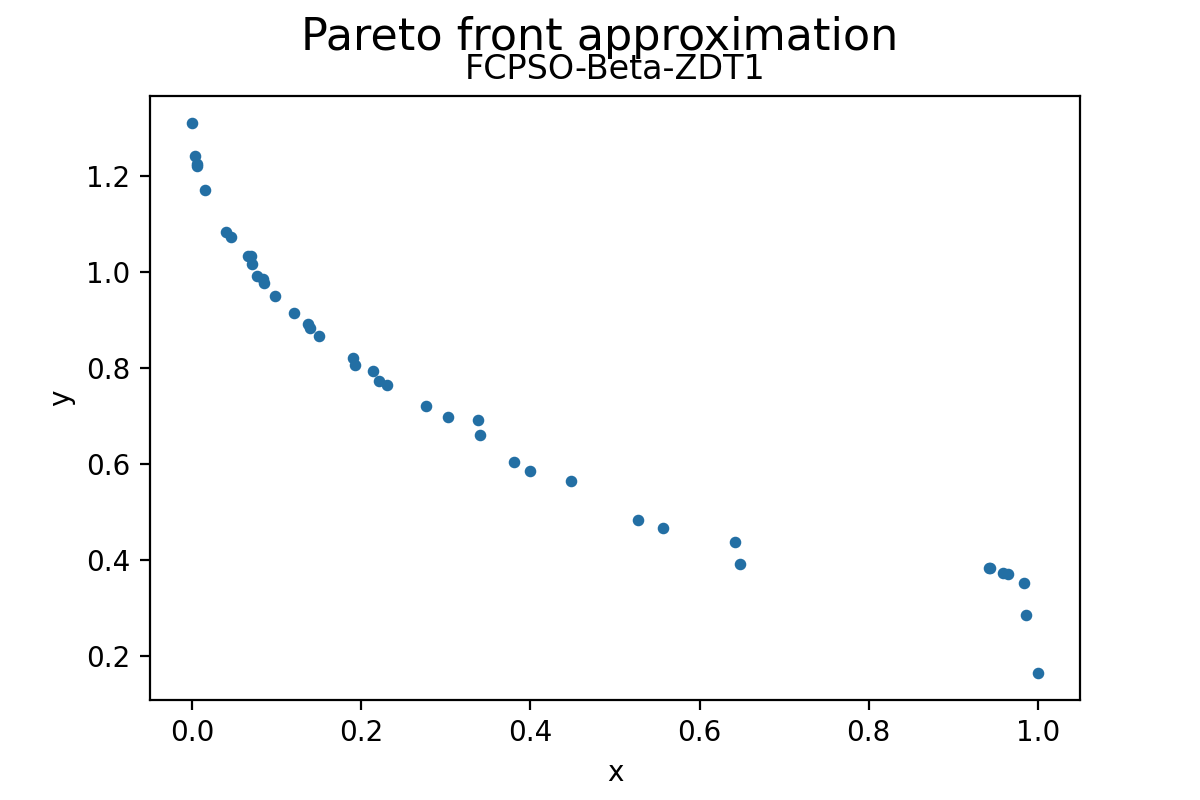}
         \caption{EM-SMPSO}
         \label{fig:emsmpso.zdt1}
     \end{subfigure}
     \caption{Pareto Fronts on ZDT1}
     \label{fig:zdt1}
\end{figure}

\begin{figure}[htbp]
     \centering
     \begin{subfigure}[b]{0.24\textwidth}
         \centering
         \includegraphics[width=\textwidth]{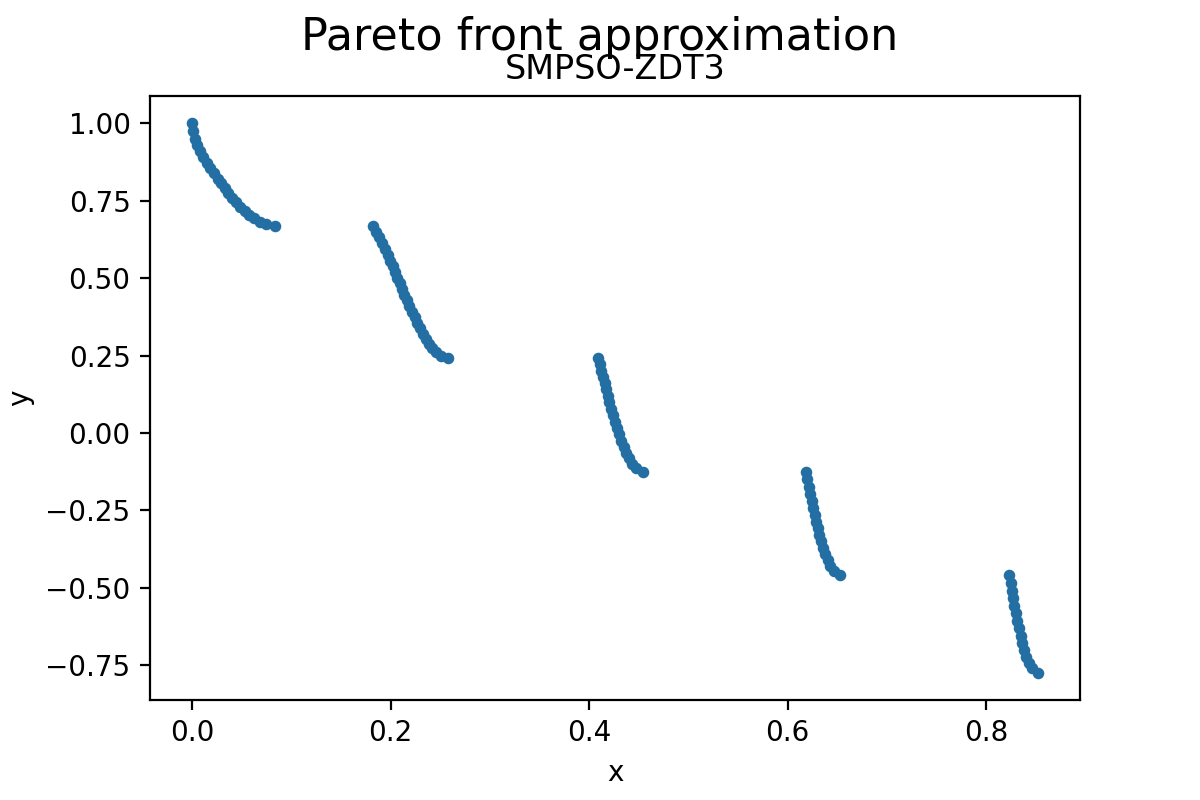}
         \caption{SMPSO}
         \label{fig:smpso.zdt3}
     \end{subfigure}
     \begin{subfigure}[b]{0.24\textwidth}
         \centering
         \includegraphics[width=\textwidth]{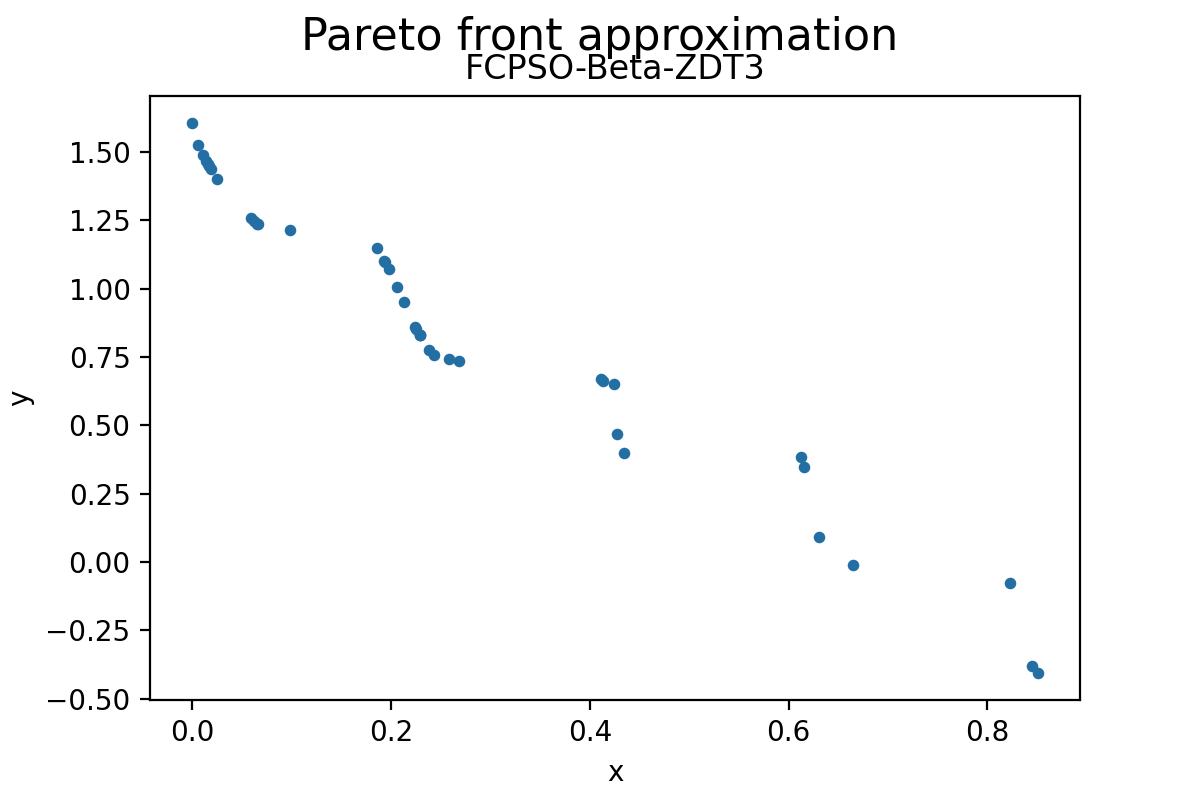}
         \caption{EM-SMPSO}
         \label{fig:emsmpso.zdt3}
     \end{subfigure}
     \caption{Pareto Fronts on ZDT3}
     \label{fig:zdt3}
\end{figure}

In figures (\ref{fig:zdt1}, \ref{fig:zdt3}), we present the Pareto fronts of EM-SMPSO (algorithm \ref{alg:emsmpso.computespeed}) on the ZDT\cite{zdt} bi-objective problems. The nature of the fronts are poor compared to that obtained by SMPSO \emph{i.e.,} significantly fewer points in the external archive and a fragmented Pareto front. The SMPSO Pareto fronts, on the other hand, are smooth and dense. The Pareto fronts were obtained using the \texttt{jmetalpy}\cite{jmetalpy} framework. 

In the single-objective realm, a blanket introduction of EM into the swarm dynamics significantly improved performance compared to vanilla PSO across various objective functions. Whereas, in the multi-objective case, that is not the case as is demonstrated by the Pareto fronts. 

\subsection{The Notion of Constriction Fairness}

It is instructive to analyse that component of SMPSO which is pivotal to its superior performance --- the constriction factor. Drawing $c_1,c_2\sim U(1.5, 2.5)$ entails that $\phi\sim U(3, 5)$. The midpoint of this distribution is $\phi=4$, which is also the value at which the two separate branches of $\chisup{v}(\phi)$ are defined in eq (\ref{eq:constriction.pso}). We say that the constriction factor is \emph{active} if the first branch is taken. Hence in the entire \emph{evolution}\footnote{
One step in the \emph{evolution} of the swarm is one iteration of the \textbf{while} loop in algorithm \ref{alg:smpso}. The complete evolution is iterating through the loop until the stopping criteria is met. 
} of the swarm, the constriction factor is activated with probability $\frac{1}{2}$. It is in this sense that SMPSO is a \emph{fairly} constricted algorithm --- the constriction factor is activated/unactivated with equal chance.

It turns out that EM-SMPSO with $\phi\sim U(3,5)$ and $\beta\sim U(0,1)$ is not a fairly constricted because of the way $\chisup{m}(\phi,\beta)$ is defined. We prove this fact in section \ref{sec:suboptimal} 

\section{Finding Parameters Schemes For Fair Constriction}

We first develop simple mathematical formulae to assess the fairness of any \emph{variant} of EM-SMPSO algorithm where $\phi\sim U(\phi_1, \phi_2)$ and $\beta\sim U(\beta_1, \beta_2)$. The respective probability densities would be
\begin{align}
    p_{\phi}(\phi) = \frac{1}{\phi_2-\phi_1} \\
    p_{\beta}(\beta) = \frac{1}{\beta_2-\beta_1}
\end{align}
Hence, these are uniform distributions. Let $\mathit{E}$ be the event that $\phi>4(1+\beta)^{-1}$ corresponding to the definition in eq (\ref{eq:constriction.empso}). We wish to find a formula for $P(E)$
\begin{align}
    P(E) &= \int\int_{\phi>4(1+\beta)^{-1}} \prob{\phi}\prob{\beta} \dd{\phi}\dd{\beta} \nonumber \\
    &= \int\int_{\beta>4\phi^{-1}-1} \prob{\beta}\prob{\phi} \dd{\beta}\dd{\phi} \label{eq:probability}
\end{align}
Using simple calculus, eq (\ref{eq:probability}) can be simplified to
\begin{align}
    P(E) = \int_{\phi_l}^{\phi_g} 
        \int_{4\phi^{-1}-1}^{\beta_2}p_\beta(\beta)p_\phi(\phi)\dd{\beta}\dd{\phi} + \int_{\phi_g}^{\phi_2}p_\phi(\phi)\dd{\phi} 
\end{align}
where
\begin{align}
    \phi_l = max(\phi_1, 4(1+\beta_2)^{-1}) \\
    \phi_g = min(4(1+\beta_1)^{-1}, \phi_2)
\end{align}
Additionally, we define the unfairness metric $\mu=P(E)-\frac{1}{2}$ to ease mathematical analysis. Note that it satisfies $-0.5\leq\mu\leq0.5$. It is a measure of how far away an algorithm is from being fairly constricted. $\mu=0$ is a fairly constricted algorithm whereas $\mu>0$ is over-constricted, and $\mu<0$ is under-constricted.

\subsection{On the suboptimal performance of EM-SMPSO}
\label{sec:suboptimal}

We have $\phi\sim U(3,5)$ and $\beta\sim U(0,1)$. Hence
\begin{align*}
    \phi_l = max(3, 2) = 3 \\
    \phi_g = min(4, 5) = 4
\end{align*}
And the probability integral with $p_\beta(\beta)=1$ and $p_\phi(\phi)=\frac{1}{2}$
\begin{align*}
    P(E) &= \int_3^4\int_{4\phi^{-1}-1}^{1}\dd{\beta}\frac{\dd{\phi}}{2} + \int_3^4\frac{\dd{\phi}}{2} \\
    &= \frac{3-4\ln(4/3)}{2}
\end{align*}
Hence, the unfairness value
\begin{align*}
    \mu &= 1-2\ln(4/3) \approx 0.42
\end{align*}
It is an over-constricted algorithm compared to SMPSO by a large margin. Thus, we are able to reason about the suboptimal nature of the Pareto Fronts using fairness analysis of the constriction factor.

\subsection{Restricted Momentum}
\label{sec:scheme1}
We wish to carry forward $\phi \sim U(3,5)$ from SMPSO. It was previously noted that using $\beta \sim U(0,1)$ caused the algorithm to be overly constricted. In other words, the algorithm used the full range of the momentum parameter. We would like to limit this range, and check if it possible to construct a fairly constricted algorithm.

For simplicity, we parameterize $\beta \sim U(0,\epsilon)$ for $0<\epsilon<1$. We restrict the full range of the momentum parameter $\beta$ by disallowing $\epsilon=1$. We wish to find the unfairness as a function of the range parameter $\epsilon$, namely $\mu(\epsilon)$.

\noindent\textbf{Choice 1} - $\epsilon<\frac{1}{3}$. We have
\begin{align}
    \phi_l &= max(3, 4(1+\epsilon)^{-1}) = 4(1+\epsilon)^{-1} \nonumber \\
    \phi_g &= min(4, 5) = 4 \nonumber
\end{align}
With $\prob{\beta}=\frac{1}{\epsilon}$ and $\prob{\phi}=\frac{1}{2}$ and hence
\begin{align}
    P(E) &= 
    \int_{4/(1+\epsilon)}^4\int_{4/\phi-1}^{\epsilon}\frac{\dd{\beta}}{\epsilon}\frac{\dd{\phi}}{2} + \int_4^5\frac{\dd{\phi}}{2}
    \nonumber \\
    &= \frac{5}{2} - \frac{2\ln(1+\epsilon)}{\epsilon}
\end{align}

\noindent\textbf{Choice 2} - $\epsilon \ge \frac{1}{3}$. We have $\phi_l=3$ and
\begin{align}
    P(E) &= 
    \int_3^4\int_{4/\phi-1}^{\epsilon}\frac{\dd{\beta}}{\epsilon}\frac{\dd{\phi}}{2} + \int_4^5\frac{\dd{\phi}}{2} \nonumber \\
    &= 1 - \frac{4\ln(4/3)-1}{2\epsilon}
\end{align}
Finally, the unfairness function
\begin{gather}
\mu(\epsilon) =
\left\{
	\begin{array}{ll}
		2\left[1 - \frac{\ln(1+\epsilon)}{\epsilon}\right]  & \mbox{ } \epsilon < \frac{1}{3} \\
		\frac{1}{2}\left[1 - \frac{4\ln(4/3)-1}{\epsilon}\right] & \mbox{ } \epsilon \geq \frac{1}{3}
	\end{array}
\right. \label{eq:unfairness.injection}
\end{gather}
\begin{figure}[]
    \centering
    \includegraphics[width=75mm]{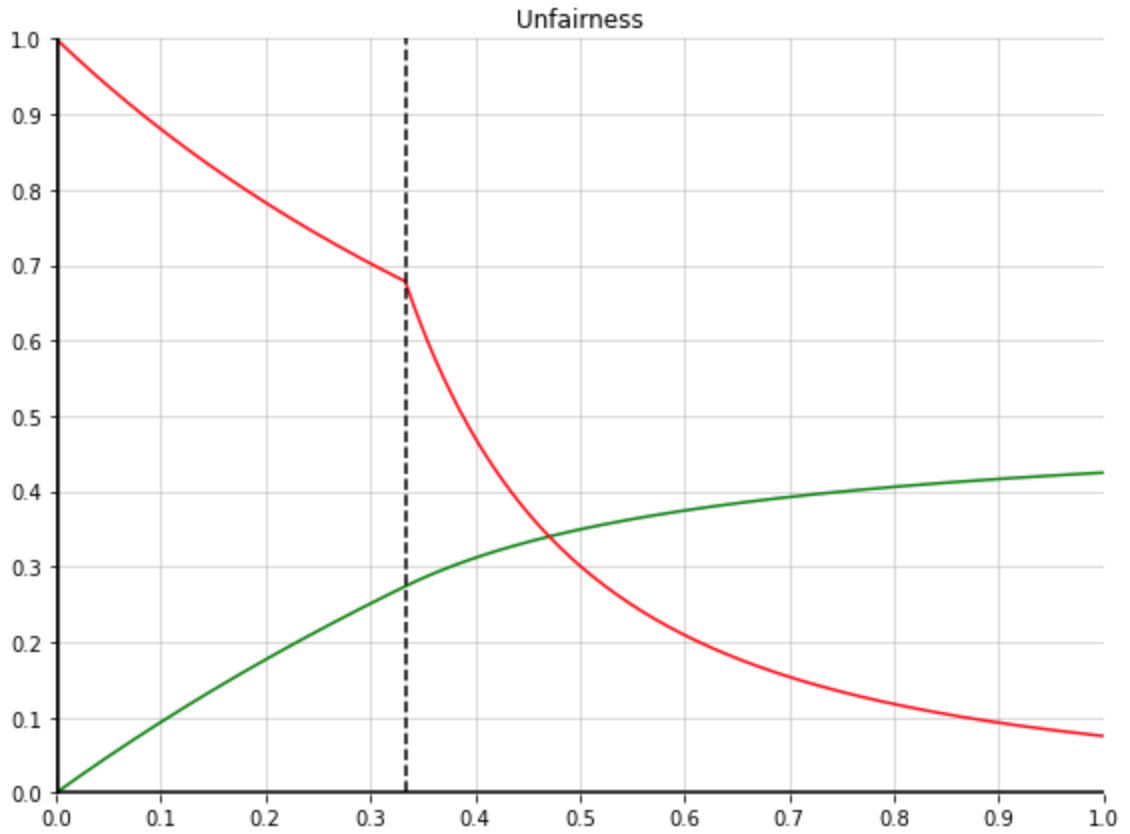}
    \caption{Green - $\mu(\epsilon)$, Red - $\dv{\mu}{\epsilon}$, Dashed - $\epsilon=\frac{1}{3}$
    }
    \label{fig:unfairness.epsilon}
\end{figure}
Is there an $\epsilon>0$ such that $\mu=0$? Consider the derivative
\begin{gather}
\dv{\mu}{\epsilon} =
\left\{
	\begin{array}{ll}
		\frac{2}{\epsilon}\left[
		\frac{\ln(1+\epsilon)}{\epsilon} - \frac{1}{1+\epsilon}
		\right]  & \mbox{ } 0 < \epsilon < \frac{1}{3} \\
		\frac{4\ln(4/3)-1}{2\epsilon^2}& \mbox{ }  \frac{1}{3} \leq \epsilon < 1
	\end{array}
\right. \label{eq:unfairness.der.injection}
\end{gather}
Clearly, $\dv{\mu}{\epsilon}>0$ for $\frac{1}{3} \leq \epsilon<1$. On the other interval, we first compute the limit (\textit{applying L'Hôpital rule twice})
\begin{align}
    \Lim{\epsilon\rightarrow0}\dv{\mu}{\epsilon} &= 1
    \label{eq:unfairness.limit}
\end{align}
Secondly, we construct the function for $0<\epsilon<\frac{1}{3}$
\begin{gather}
    g(\epsilon) = \ln(1+\epsilon) - \frac{\epsilon}{1+\epsilon}
\end{gather}
With $g'(\epsilon)=\frac{\epsilon}{(1+\epsilon)^2}>0$, and $\Lim{\epsilon\rightarrow0}g(\epsilon)=0$ 
\begin{align}
    \ln(1+\epsilon) - \frac{\epsilon}{1+\epsilon} &> 0 \nonumber \\
    \epsilon^2\dv{\mu}{\epsilon} &> 0 \nonumber \\
    \implies \dv{\mu}{\epsilon} &> 0 &(\epsilon\neq0)\label{eq:unfairness.der.sign}
\end{align}
Based on eqs (\ref{eq:unfairness.limit}, \ref{eq:unfairness.der.sign}), we can conclude that $\mu(\epsilon)\neq0$ for $0 < \epsilon < 1$. Hence, it is impossible to obtain a fairly constricted algorithm using the scheme of $\beta\sim U(0, \epsilon)$.

We have from elementary calculus that, $\Lim{\epsilon \rightarrow 0}\frac{\ln(1+\epsilon)}{\epsilon}=1$ and hence
\begin{gather}
    \Lim{\epsilon\rightarrow0}\mu(\epsilon)=0\label{eq:almost.fair}
\end{gather}
Hence, one could construct an algorithm which is \emph{almost} fairly constricted and by allowing $\epsilon$ to be infinitesimally small. However, such an algorithm would lose its ultimate purpose of being able to exploit exponentially averaged momentum because sampling infinitesimally small values of $\beta$ essentially means that momentum is absent. 
\subsection{Fairly Constricted Parameter Set}
\label{sec:fairly.set}
\begin{figure}[t]
    \centering
    \includegraphics[width=75mm]{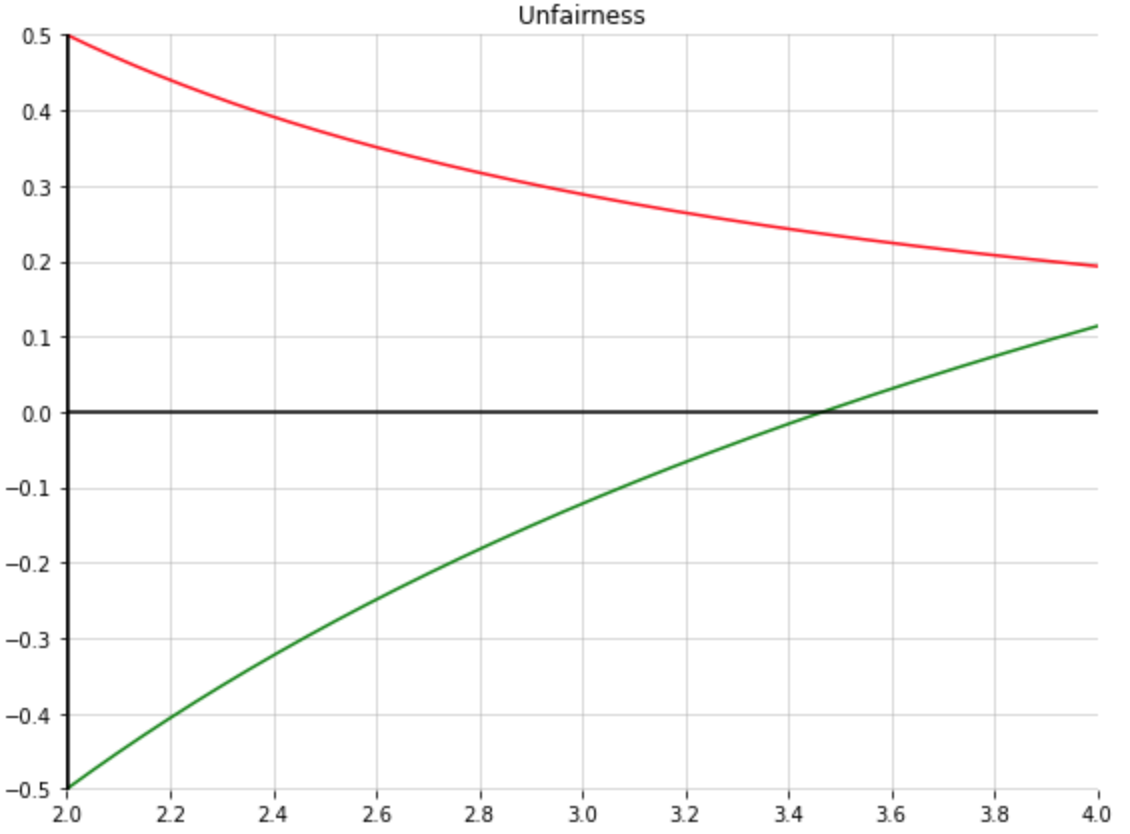}
    \caption{Green - $\mu(\phi)$, Red - $\dv{\mu}{\phi}$}
    \label{fig:unfairness.phi.2d}
\end{figure}

We wish to utilize the full range of the momentum parameter and hence set $\beta_1=0, \beta_2=1$. In computing the probability integral, we posit $\phi_l=\phi_1$ and $\phi_g=\phi_2$ which amounts to exercising the choices of $\phi_1\geq2$ and $\phi_2\leq4$ respectively. Hence
\begin{align}
    P(E) &= \int_{\phi_1}^{\phi_2}\int_{4/\phi-1}^{1}\dd{\beta}\frac{\dd{\phi}}{\phi_2-\phi_1} \nonumber \\
    &= 2 - 4\frac{\ln(\phi_2/\phi_1)}{\phi_2-\phi_1} \label{eq:phi.2d}
\end{align}
\begin{figure}[b]
    \centering
    \includegraphics[width=75mm]{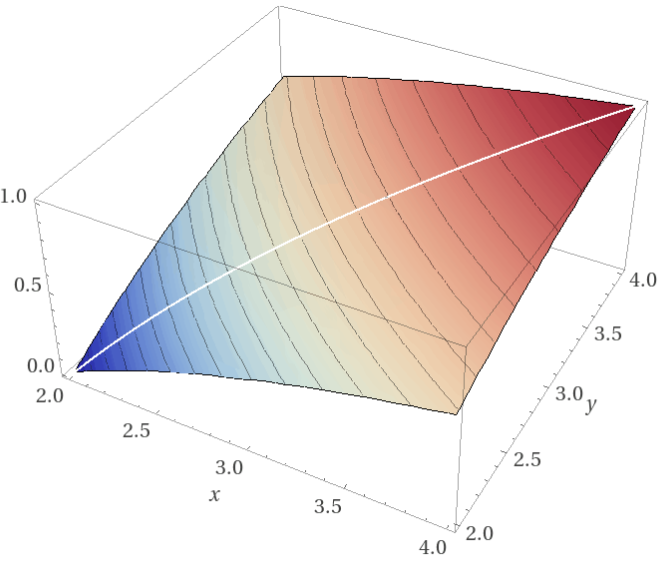}
    \caption{$X\text{-axis}\rightarrow\phi_1$, $Y\text{-axis}\rightarrow\phi_2$,
    $Z\text{-axis}\rightarrow P(E)$}
    \label{fig:unfairness.phi.3d}
\end{figure}
Let us arbitrarily assign $\phi_1=2$. This is reasonable as per the previous parameter sets studied. We obtain an the unfairness as a function of $\phi_2$
\begin{gather}
    \mu(\phi_2) = \frac{3}{2} - 4\frac{\ln(\phi_2/2)}{\phi_2-2} \label{eq:phi.1d}
\end{gather}
We transform $\frac{\phi_2}{2} \rightarrow x$ and set $\mu(x)=0$ to obtain the following transcendental equation in the variable $x$
\begin{gather}
    \psi(x)\equiv\frac{x-1}{\ln x} - \frac{4}{3} = 0
\end{gather}
where $\psi(x)$ has been defined for convenience. A solution $\bar{x}$ to this equation must lie in $1<\bar{x}\leq2$. Note that $\Lim{x\rightarrow1}\psi(x)=\frac{-1}{3}<0$ and $\psi(2)=\frac{1}{\ln2} - \frac{4}{3}>0$. It is well known that $\psi(x)$ is monotonically increasing (\emph{it is of the form of the asymptotic prime counting function} \cite{prime}) and thus a unique solution exists in the range $(1, 2]$. This can also be visually confirmed from the plot of eq (\ref{eq:phi.1d}) in figure (\ref{fig:unfairness.phi.2d}) \footnote{
In the plot, the independent variable of the X-axis is the same as $\phi_2$ from eq \ref{eq:phi.1d}. In other words, $\phi \equiv \phi_2$.
}

Wolfram Alpha \cite{wolfram} outputs the solution as $\bar{x}\approx1.7336$ and we obtain $\phi_2=2\bar{x}\approx3.4672$. Hence using $c_1,c_2\sim U(1, 1.7336)$ and $\beta\sim U(0,1)$ would result in a fairly constricted algorithm. We call it \emph{Fairly Constricted Particle Swarm Optimization} (\textbf{FCPSO}).

The surface plot of eq (\ref{eq:phi.2d}) in the $(\phi_1,\phi_2)$ plane is plotted in figure (\ref{fig:unfairness.phi.3d}). Note that there may exist other parameter sets that are also fairly constricting. In this work, we have derived only one fairly constricting set and subsequently used it for benchmarking. 

\section{Results}

\subsection{Pareto Fronts}
We first present the Pareto fronts of the ZDT1 and ZDT3 problems (\emph{first shown in Section \ref{sec:prelim}}). 
\begin{figure}[htbp]
     \centering
     \begin{subfigure}[b]{0.24\textwidth}
         \centering
         \includegraphics[width=\textwidth]{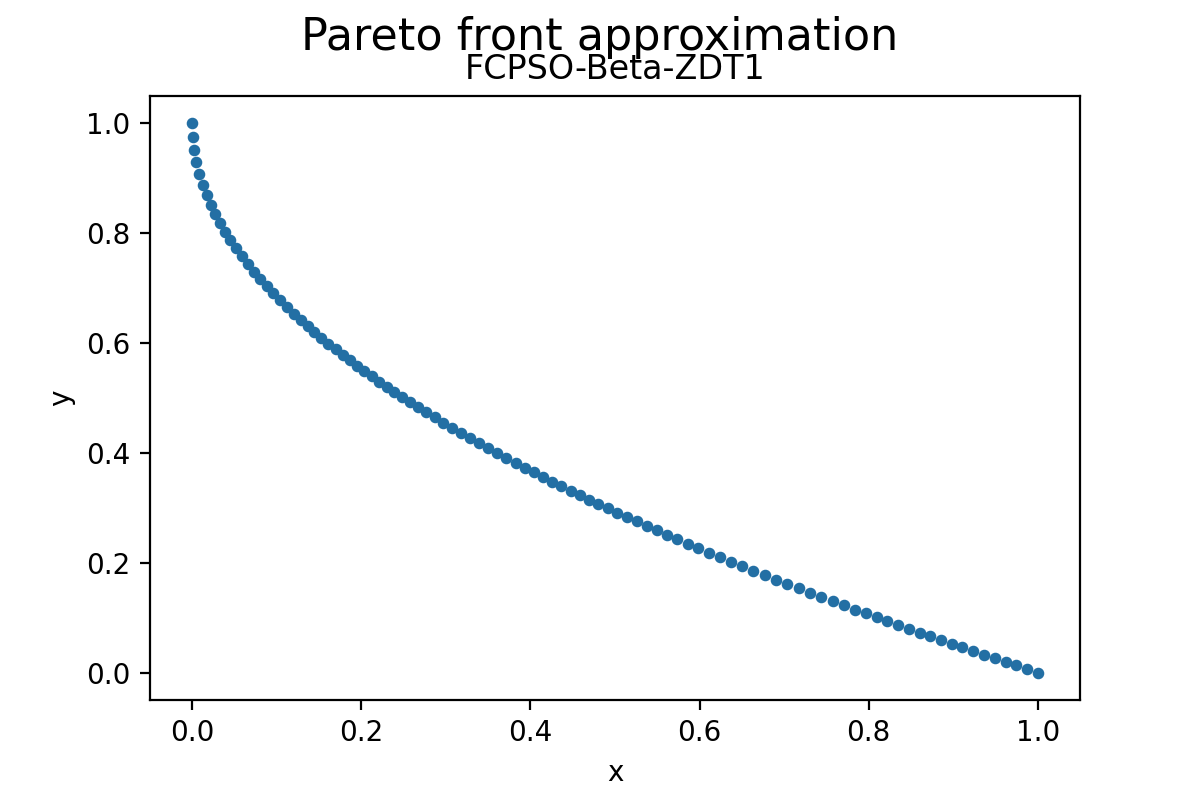}
         \caption{FCPSO on ZDT1}
         \label{fig:fcpso.zdt1}
     \end{subfigure}
     \begin{subfigure}[b]{0.24\textwidth}
         \centering
         \includegraphics[width=\textwidth]{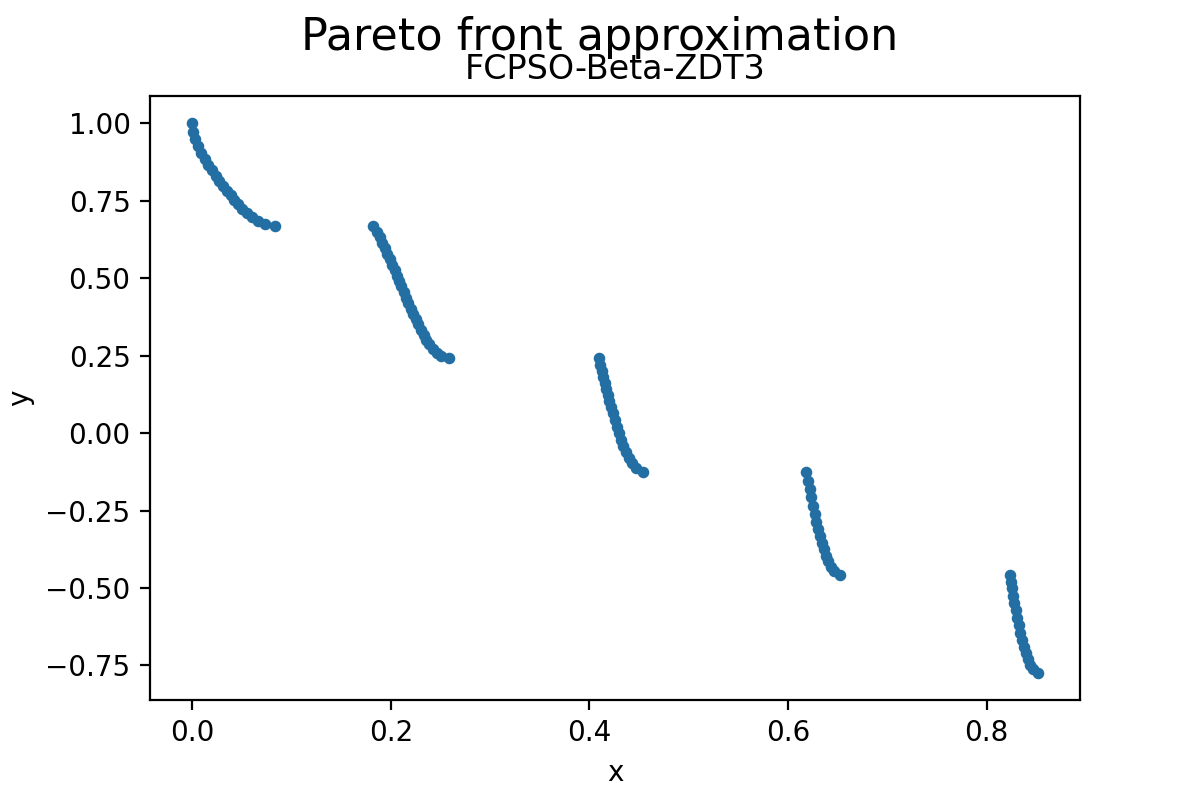}
         \caption{FCPSO on ZDT3}
         \label{fig:fcpso.zdt3}
     \end{subfigure}
     \caption{Pareto Fronts of FCPSO}
     \label{fig:fcpso.fronts}
\end{figure}

From a first qualitative look, the Pareto fronts of FCPSO match that of SMPSO. The solution points are densely packed, and well-connected unlike the fragmented Pareto fronts of the naive EM-SMPSO algorithm. 

\subsection{Assessment with Quality Indicators}

We choose the following 5 quality indicators to assess the performance of FCPSO. Namely, they are Inverted Generational Distance (IGD), Spacing (SP), Hypervolume (HV), and the $\epsilon$-indicator (EPS). A thorough description of these indicators can be found in \cite{audet}. 

The measurement of all indicators was done after letting the swarm evolve for $25,000$ function evaluations. In the case of measuring function evaluation itself, we allow the swarm to evolve until $95\%$ of the hypervolume (HV hereafter) of the theoretically computed Pareto front is reached. The theoretical fronts were obtained from \cite{coello.handbook, optproblems} and \cite{moea.framework}. 

All quality indicator values of FCPSO are accompanied by corresponding values from SMPSO for the sake of comparison. Each measurement was repeated $20$ times for statistical testing. The resultant $p$-values have been written as subscripts in the tables. We have not shown values of all quality indicators for all problems due to space constraints, however. 

\subsubsection{Bi-objective ZDT and Tri-objective DTLZ}

\begin{table}[h]
\begin{tabular}{|c|c|c|}
\hline
\textbf{FE}      & \textbf{SMPSO} & \textbf{FCPSO} \\ \hline
\textbf{zdt1} &  72.12  &          \textbf{56.45}\sub{5.65e-10}  \\ \hline
\textbf{zdt2} &  \textbf{68.95}  &          87.90\sub{1.10e-01}  \\ \hline
\textbf{zdt3} & 110.90  &         \textbf{102.01}\sub{5.11e-03}  \\ \hline
\textbf{zdt4} &  \textbf{32.92}  &          45.35\sub{5.73e-07}  \\ \hline
\textbf{zdt6} &  \textbf{27.40}  &          29.31\sub{9.20e-01}  \\ \hline
\textbf{dtlz1} &  \textbf{39.49}  &          58.40\sub{1.37e-03}  \\ \hline
\textbf{dtlz2} &   2.16  &           2.19 \\ \hline
\textbf{dtlz3} & 521.41  &         \textbf{402.91}\sub{1.59e-06} \\ \hline
\textbf{dtlz4} &   \textbf{9.31}  &          10.52 \\ \hline
\textbf{dtlz5} &   6.67 &           8.68  \\ \hline
\textbf{dtlz6} & \textbf{107.27}  &         175.70\sub{1.06e-09} \\ \hline
\textbf{dtlz7} & 102.15 &          \textbf{74.30}\sub{1.16e-07} \\ \hline
\end{tabular}%
\begin{tabular}{|c|c|c|}
\hline
\textbf{HV}      & \textbf{SMPSO} & \textbf{FCPSO} \\ \hline
\textbf{zdt1} &   3.66 &           3.66\subzero \\ \hline
\textbf{zdt2} &   3.33 &           3.33\subzero \\ \hline
\textbf{zdt3} &   \textbf{4.40} &           4.38\sub{4.11e-11} \\ \hline
\textbf{zdt4} &   3.65 &           3.65\subzero \\ \hline
\textbf{zdt6} &   3.17 &           3.17\sub{1.55e-10} \\ \hline
\textbf{dtlz1} &   \textbf{3.28} &           3.27\sub{2.70e-03} \\ \hline
\textbf{dtlz2} &   \textbf{7.34} &           7.33\sub{6.02e-13} \\ \hline
\textbf{dtlz3} &   \textbf{6.39} &           5.97\sub{2.70e-03} \\ \hline
\textbf{dtlz4} &   7.33 &           7.29\subzero \\ \hline
\textbf{dtlz5} &   4.26 &           4.26\subzero \\ \hline
\textbf{dtlz6} &   \textbf{4.26} &           3.79\sub{3.18e-04} \\ \hline
\textbf{dtlz7} &  \textbf{11.07} &          11.01\sub{6.33e-05} \\ \hline
\end{tabular}
\caption{ZDT \& DTLZ : FE and HV}
\label{table:2zdt.3dtlz}
\end{table}

The ZDT \cite{zdt} suite of multi-objective problems is a list of biobjective problems for the evaluation of MOO algorithms. Along with the triobjective DTLZ \cite{dtlz} problems, these constitute the basic problems that any MOO algorithm must be able to solve accurately. We show that FCPSO matches the performance of the state-of-the-art SMPSO in these problems. Please refer to table (\ref{table:2zdt.3dtlz}) for the FE and HV measurements. FCPSO matches SMPSO in most problems, occasionally outperforming (\emph{and underperforming}) SMPSO.

A peculiarity to be noted for the DTLZ-2, 4 and 5 problems is that the number of FEs is not characteristic of the other problems and hence we have not included their statistical p-values. This is probably due to the particular nature of the problem, or the swarm initialisation implemented in \texttt{jmetalpy}.

\subsubsection{5-objective and 10-objective DTLZ}

We test the algorithm on a harder variant of the DTLZ problems with 5 objectives. The quality indicator values are shown in tables \ref{table:5dtlz.hv.igd} and \ref{table:5dtlz.eps.sp}. For 10-objective DTLZ, we do not have HV values as \texttt{jmetalpy} took too long to evaluate HV values. Thus, we only have IGD, EPS and SP values pertaining to these problems. The values are available in table \ref{table:10dtlz.igd.eps} and \ref{table:10dtlz.sp}

\begin{table}[h]
\begin{tabular}{|c|c|c|}
\hline
\textbf{HV}      & \textbf{SMPSO} & \textbf{FCPSO} \\ \hline
\textbf{dtlz1} &   1.73 &  \textbf{6.30}\subzero \\ \hline
\textbf{dtlz2} &  22.03 &  \textbf{27.56}\subzero \\ \hline
\textbf{dtlz3} &   0.24 &  \textbf{5.34}\sub{3.80e-08} \\ \hline
\textbf{dtlz4} &  30.45 &  \textbf{30.85}\sub{1.99e-07} \\ \hline
\textbf{dtlz5} &   6.11 &  \textbf{6.15}\sub{1.99e-07} \\ \hline
\textbf{dtlz6} &   0.11 &  \textbf{6.02}\subzero \\ \hline
\textbf{dtlz7} &  16.89 &  \textbf{54.24}\subzero \\ \hline
\end{tabular}%
\begin{tabular}{|c|c|c|}
\hline
\textbf{IGD}      & \textbf{SMPSO} & \textbf{FCPSO} \\ \hline
\textbf{dtlz1} &   4.36 & \textbf{0.07}\subzero \\ \hline
\textbf{dtlz2} &   0.64 & \textbf{0.42}\subzero \\ \hline
\textbf{dtlz3} &  50.77 & \textbf{8.16}\sub{6.22e-15} \\ \hline
\textbf{dtlz4} &   0.28 & \textbf{0.18}\subzero \\ \hline
\textbf{dtlz5} &   0.07 & \textbf{0.06}\sub{1.62e-01} \\ \hline
\textbf{dtlz6} &   2.83 & \textbf{0.11}\subzero \\ \hline
\textbf{dtlz7} &   1.12 & \textbf{0.50}\subzero \\ \hline
\end{tabular}
\caption{5-DTLZ : HV and IGD}
\label{table:5dtlz.hv.igd}
\end{table}%
\begin{table}[h]
\begin{tabular}{|c|c|c|}
\hline
\textbf{EPS}      & \textbf{SMPSO} & \textbf{FCPSO} \\ \hline
\textbf{dtlz1} &   3.37  &           \textbf{0.12}\subzero \\ \hline
\textbf{dtlz2} &   0.71  &           \textbf{0.55}\subzero \\ \hline
\textbf{dtlz3} &  36.81  &           \textbf{6.19}\sub{2.95e-14} \\ \hline
\textbf{dtlz4} &   0.43  &           \textbf{0.32}\sub{6.66e-08} \\ \hline
\textbf{dtlz5} &   0.09  &           \textbf{0.08}\sub{1.62e-01} \\ \hline
\textbf{dtlz6} &   2.96  &           \textbf{0.11}\subzero \\ \hline
\textbf{dtlz7} &   2.77  &           \textbf{1.05}\sub{4.11e-11} \\ \hline
\end{tabular}%
\begin{tabular}{|c|c|c|}
\hline
\textbf{SP}      & \textbf{SMPSO} & \textbf{FCPSO} \\ \hline
\textbf{dtlz1} &   3.10  &           \textbf{2.67}\sub{1.37e-03} \\ \hline
\textbf{dtlz2} &   \textbf{0.29} &           0.34\subzero \\ \hline
\textbf{dtlz3} &  49.05  &          \textbf{39.85}\sub{4.11e-11} \\ \hline
\textbf{dtlz4} &   \textbf{0.18}  &           0.21\sub{1.59e-06} \\ \hline
\textbf{dtlz5} &   \textbf{0.20}  &           0.25 \subzero \\ \hline
\textbf{dtlz6} &   1.11  &           \textbf{0.56}\subzero \\ \hline
\textbf{dtlz7} &   0.25  &           \textbf{0.23}\sub{2.70e-03} \\ \hline
\end{tabular}
\caption{5-DTLZ : EPS and SP}
\label{table:5dtlz.eps.sp}
\end{table}

FCPSO outperforms SMPSO in all problems except DTLZ2, DTLZ4, DTLZ5 with respect to the spacing (SP) quality indicator in both 5-objective and 10-objective realm. There is a notable exception, however, where SMPSO dominates with respect to SP in 10-objectives. Nevertheless, the gap between it and FCPSO is not significantly high.

\begin{table}[h]
\begin{tabular}{|c|c|c|}
\hline
\textbf{IGD}      & \textbf{SMPSO} & \textbf{FCPSO} \\ \hline
\textbf{dtlz1} &   8.09  &           \textbf{1.93}\sub{1.25e-12} \\ \hline
\textbf{dtlz2} &   0.75  &           \textbf{0.57}\subzero \\ \hline
\textbf{dtlz3} &  43.87  &          \textbf{31.48}\sub{1.37e-03} \\ \hline
\textbf{dtlz4} &   0.58  &           \textbf{0.43}\subzero \\ \hline
\textbf{dtlz5} &   \textbf{0.06}  &           0.08\sub{1.97e-09} \\ \hline
\textbf{dtlz6} &   0.51  &           \textbf{0.15}\sub{1.97e-09} \\ \hline
\textbf{dtlz7} &   1.45  &           \textbf{1.32}\sub{4.22e-06} \\ \hline
\end{tabular}%
\begin{tabular}{|c|c|c|}
\hline
\textbf{EPS}      & \textbf{SMPSO} & \textbf{FCPSO} \\ \hline
\textbf{dtlz1} &   5.61  &           \textbf{1.57}\sub{2.98e-10} \\ \hline
\textbf{dtlz2} &   0.65  &           \textbf{0.56}\subzero \\ \hline
\textbf{dtlz3} &  33.16  &          \textbf{21.56}\sub{1.45e-04} \\ \hline
\textbf{dtlz4} &   0.65  &           \textbf{0.52}\subzero \\ \hline
\textbf{dtlz5} &   \textbf{0.08}  &           0.10\sub{6.33e-05} \\ \hline
\textbf{dtlz6} &   0.62  &           \textbf{0.15}\sub{2.56e-12} \\ \hline
\textbf{dtlz7} &   1.54  &           \textbf{0.95}\sub{1.64e-02} \\ \hline
\end{tabular}
\caption{10-DTLZ : IGD and EPS}
\label{table:10dtlz.igd.eps}
\end{table}
\begin{table}[h]
\centering
\begin{tabular}{|c|c|c|}
\hline
\textbf{SP}      & \textbf{SMPSO} & \textbf{FCPSO} \\ \hline
\textbf{dtlz1} &  13.92 &          \textbf{10.29}\sub{1.55e-10} \\ \hline
\textbf{dtlz2} &   \textbf{0.44} &           0.62\subzero \\ \hline
\textbf{dtlz3} &  \textbf{90.24} &          93.83\sub{7.19e-02} \\ \hline
\textbf{dtlz4} &   \textbf{0.43} &           0.44\sub{1.10e-01} \\ \hline
\textbf{dtlz5} &   \textbf{0.22} &           0.30\subzero \\ \hline
\textbf{dtlz6} &   \textbf{0.80} &           0.95\sub{1.33e-15} \\ \hline
\textbf{dtlz7} &   0.92 &           \textbf{0.74}\sub{6.02e-13} \\ \hline
\end{tabular}
\caption{10-DTLZ : SP}
\label{table:10dtlz.sp}
\end{table}

\subsubsection{5-objective and 10-objective WFG}

The WFG test suite was proposed in \cite{husband} to overcome the limitations of ZDT/DTLZ test suites. For one, ZDT is limited to 2 objectives only. Secondly, the DTLZ problems are not \emph{deceptive} (\emph{a notion developed in \cite{husband}}) and none of them feature a large flat landscape. Moreover, they state that the nature of the Pareto-front for DTLZ-5,6 is unclear beyond 3 objectives. Lastly, the complexity of each of the previous mentioned problem is \emph{fixed} for a particular problem. Hence, the WFG problems are expressed as a generalised scheme of transformations that lead an input vector to a point in the objective space. The WFG test suite is harder and a rigorous attempt at creating an infallible, robust benchmark for MOO solvers.

\begin{table}[h]
\centering
\begin{tabular}{|c|c|c|}
\hline
\textbf{}      & \textbf{SMPSO} & \textbf{FCPSO} \\ \hline
\textbf{wfg1} &   1.21  &           1.39\subzero \\ \hline
\textbf{wfg2} &   0.12  &           0.14\sub{3.18e-04} \\ \hline
\textbf{wfg3} &   \textbf{0.09}  &           0.12\subzero \\ \hline
\textbf{wfg4} &   \textbf{1.93}  &           2.27\sub{2.22e-16} \\ \hline
\textbf{wfg5} &   \textbf{0.75}  &           0.91\subzero \\ \hline
\textbf{wfg6} &   \textbf{0.23}  &           0.27\sub{5.73e-07} \\ \hline
\textbf{wfg7} &   2.51 &           2.48\sub{1.62e-01} \\ \hline
\textbf{wfg8} &   \textbf{0.23}  &           0.27\sub{6.63e-09} \\ \hline
\textbf{wfg9} &   2.06  &           \textbf{1.43}\subzero \\ \hline
\end{tabular}
\caption{10-WFG : SP}
\label{table:10wfg.sp}
\end{table}
\begin{table}[h]
\begin{tabular}{|c|c|c|}
\hline
\textbf{HV}      & \textbf{SMPSO} & \textbf{FCPSO} \\ \hline
\textbf{wfg1} & 4450.24  &        \textbf{4455.31}\subzero \\ \hline
\textbf{wfg2} & 1693.23  &        \textbf{1987.07}\sub{2.56e-12} \\ \hline
\textbf{wfg3} &   14.51 &          \textbf{14.59}\sub{3.18e-04} \\ \hline
\textbf{wfg4} & 1395.45 &        \textbf{1604.41}\subzero \\ \hline
\textbf{wfg5} & 2700.61 &        \textbf{2813.04}\subzero \\ \hline
\textbf{wfg6} & 1540.74 &        \textbf{1557.22}\sub{2.95e-14} \\ \hline
\textbf{wfg7} & 1909.03  &        \textbf{1986.85}\sub{1.33e-15} \\ \hline
\textbf{wfg8} & 1910.25 &        \textbf{1886.20}\sub{4.22e-06} \\ \hline
\textbf{wfg9} & \textbf{2348.75} &        2237.12\sub{9.32e-03} \\ \hline
\end{tabular}%
\begin{tabular}{|c|c|c|}
\hline
\textbf{IGD}      & \textbf{SMPSO} & \textbf{FCPSO} \\ \hline
\textbf{wfg1} &   \textbf{1.86}  &           1.88\sub{1.36e-13} \\ \hline
\textbf{wfg2} &   1.95  &           \textbf{1.75}\subzero \\ \hline
\textbf{wfg3} &   2.63  &           \textbf{2.61}\sub{2.14e-08} \\ \hline
\textbf{wfg4} &   1.74  &           \textbf{1.69}\sub{2.56e-12} \\ \hline
\textbf{wfg5} &   1.42  &           \textbf{1.38}\subzero \\ \hline
\textbf{wfg6} &   3.13  &           \textbf{3.11}\sub{2.22e-16} \\ \hline
\textbf{wfg7} &   2.04  &           \textbf{2.00}\sub{2.56e-12} \\ \hline
\textbf{wfg8} &   2.93  &           \textbf{2.92}\subzero \\ \hline
\textbf{wfg9} &   1.52  &           \textbf{1.50}\sub{2.78e-02} \\ \hline
\end{tabular}
\caption{5-WFG : HV and IGD}
\label{table:5wfg.hv.igd}
\end{table}
\begin{table}
\begin{tabular}{|c|c|c|}
\hline
\textbf{EPS}      & \textbf{SMPSO} & \textbf{FCPSO} \\ \hline
\textbf{wfg1} &   \textbf{1.55}  &           1.65\subzero \\ \hline
\textbf{wfg2} &   7.96  &           \textbf{7.38}\sub{1.08e-05} \\ \hline
\textbf{wfg3} &   6.83  &           6.83\sub{5.49e-01} \\ \hline
\textbf{wfg4} &   2.19  &           \textbf{2.04}\sub{1.59e-06} \\ \hline
\textbf{wfg5} &   \textbf{5.00}  &           5.01\sub{3.17e-01} \\ \hline
\textbf{wfg6} &   7.44  &           \textbf{7.43}\sub{6.89e-01} \\ \hline
\textbf{wfg7} &   2.45  &           \textbf{2.40}\sub{9.32e-03} \\ \hline
\textbf{wfg8} &   \textbf{8.32}  &           8.33\sub{1.36e-13} \\ \hline
\textbf{wfg9} &   \textbf{2.36}  &           3.58\subzero \\ \hline
\end{tabular}%
\begin{tabular}{|c|c|c|}
\hline
\textbf{SP}      & \textbf{SMPSO} & \textbf{FCPSO} \\ \hline
\textbf{wfg1} &   0.43 &           0.44\sub{2.30e-01} \\ \hline
\textbf{wfg2} &   \textbf{0.04} &           0.06\sub{6.66e-08} \\ \hline
\textbf{wfg3} &   \textbf{0.04} &           0.05\subzero \\ \hline
\textbf{wfg4} &   \textbf{0.58} &           0.69\sub{2.22e-16} \\ \hline
\textbf{wfg5} &   \textbf{0.28} &           0.31\sub{5.73e-07} \\ \hline
\textbf{wfg6} &   \textbf{0.09} &           0.11\sub{1.97e-09} \\ \hline
\textbf{wfg7} &   0.63 &           \textbf{0.62}\sub{8.41e-01} \\ \hline
\textbf{wfg8} &   \textbf{0.07} &           0.08\sub{5.73e-07} \\ \hline
\textbf{wfg9} &   0.64 &           \textbf{0.60}\sub{4.22e-06} \\ \hline
\end{tabular}
\caption{5-WFG : EPS and SP}
\label{table:5wfg.eps.sp}
\end{table}

\begin{table}[h]
\begin{tabular}{|c|c|c|}
\hline
\textbf{IGD}      & \textbf{SMPSO} & \textbf{FCPSO} \\ \hline
\textbf{wfg1} &   3.29  &           \textbf{3.26}\sub{5.49e-01} \\ \hline
\textbf{wfg2} &   6.14  &           \textbf{5.86}\sub{5.73e-07} \\ \hline
\textbf{wfg3} &   5.73  &           \textbf{5.70}\sub{1.97e-09} \\ \hline
\textbf{wfg4} &   7.09  &           \textbf{5.30}\sub{6.22e-15} \\ \hline
\textbf{wfg5} &   0.17  &           \textbf{0.16}\subzero \\ \hline
\textbf{wfg6} &  13.10  &          \textbf{13.07}\subzero \\ \hline
\textbf{wfg7} &   5.17  &           \textbf{4.89}\sub{1.05e-11} \\ \hline
\textbf{wfg8} &   \textbf{8.59}  &           8.60\sub{5.49e-01} \\ \hline
\textbf{wfg9} &   2.48  &           \textbf{2.12}\subzero \\ \hline
\end{tabular}%
\begin{tabular}{|c|c|c|}
\hline
\textbf{EPS}      & \textbf{SMPSO} & \textbf{FCPSO} \\ \hline
\textbf{wfg1} &   \textbf{1.48} &           1.52\sub{2.78e-02} \\ \hline
\textbf{wfg2} &  15.12 &          \textbf{14.22}\sub{2.67e-05} \\ \hline
\textbf{wfg3} &  13.13 &          \textbf{13.12}\sub{2.78e-02} \\ \hline
\textbf{wfg4} &   3.78 &           \textbf{2.98}\sub{5.65e-10} \\ \hline
\textbf{wfg5} &   0.06 &           \textbf{0.05}\subzero \\ \hline
\textbf{wfg6} &  14.77 &          \textbf{14.76}\sub{8.91e-02} \\ \hline
\textbf{wfg7} &   \textbf{4.24} &           4.27\sub{2.30e-01} \\ \hline
\textbf{wfg8} &  \textbf{16.63} &          16.64\subzero \\ \hline
\textbf{wfg9} &   1.16 &           \textbf{0.69}\subzero \\ \hline
\end{tabular}
\caption{10-WFG : IGD and EPS}
\label{table:10wfg.igd.eps}
\end{table}%

Tables \ref{table:5wfg.hv.igd} and \ref{table:5wfg.eps.sp} contain the results for 5-objective WFG problems. The results for 10-objective WFG problems are in tables \ref{table:10wfg.igd.eps} and \ref{table:10wfg.sp}. FCPSO matches SMPSO with a small margin in most problems, if not outperforming it.

\subsection{The Effect of Unfairness on Performance}

\begin{figure}[htbp]
     \centering
     \begin{subfigure}[b]{0.35\textwidth}
         \centering
         \includegraphics[width=\textwidth]{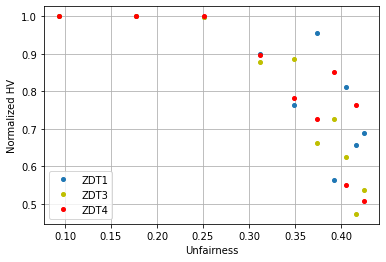}
     \end{subfigure}
     
     \begin{subfigure}[b]{0.35\textwidth}
         \centering
         \includegraphics[width=\textwidth]{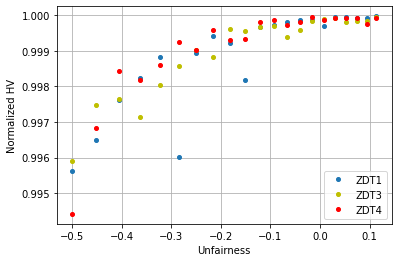}
     \end{subfigure}
     \caption{Normalized HV vs. Unfairness}
     \label{fig:unf}
\end{figure}

The unfairness factor $\mu$ is central to the performance of FCPSO. This is evident from the drastic difference in the nature of the Pareto fronts between the naive EM-SMPSO (figure \ref{fig:emsmpso.zdt1}) and FCPSO (figure \ref{fig:fcpso.zdt1}) algorithm. However, it is informative to see how the performance varies for other values of unfairness (apart from $\mu=0$). The range of unfairness values exhibited by the parameter sets discussed in this work are as follows
\begin{itemize}
    \item Eq \ref{eq:unfairness.injection} - $[0, 0.424]$
    \item Eq \ref{eq:phi.1d} - $(-0.5, 0.113]$
\end{itemize}
Despite the full range of unfairness $[-0.5, 0.5]$ not being covered, we go over a wide enough range in both under-constricted and over-constricted regions. We profile ZDT1, ZDT3 and ZDT4 in these ranges and assess the performance by computing HV (figure \ref{fig:unf}). The plotted HV values have been normalized to the HV obtained by running vanilla SMPSO.

In the over-constricted range, performance degrades significantly beyond $\mu=0.25$ dropping to half of the optimal HV. For $0.1\leq\mu\leq0.25$, the HV is stable with respect to the unfairness. In the under-constricted region, however, the HV does not go below $99.5\%$ of the optimal HV. Hence, over-constriction appears to affect the performance more than under-constriction. Moreover, the trend in the HV values with respect to unfairness is consistent among all the three problems considered.

\section{Discussion and Impact of our Work}
\label{sec:discussion}

At the time of appearance, SMPSO was the state-of-the-art MOO solver compared to other algorithms such as OMOPSO, NSGA-II. Its success is tied to the use of velocity constriction, which we have theoretically analysed and extended to the case of exponentially-averaged momentum. Moreover, there is a dearth of literature on the stochastic analysis of evolutionary algorithms. In the realm of single-objective PSO, \cite{jiang} has analysed the stability of PSO considering the stochastic nature of $r_1, r_2$ of the PSO update equations \cite{kennedy.eberhart.og}. We have successfully performed an analysis in a similar vein. The idea proposed in this work is simple, but it could be applied for the stochastic analysis of evolutionary algorithms.

\section{Conclusion and Future Works}
\label{sec:conclusion}

In this paper, we have discussed the motivations for introducing exponentially-averaged momentum in the SMPSO framework. Having defined specific notions for constriction fairness,  we have successfully incorporated exponentially-averaged momentum to SMPSO and demonstrated its performance in MOO problems. It would be beneficial to develop a large number of parameter schemes that are also fairly constricting and compare their performance. Finding a parameterization $(\phi_1, \phi_2, \beta_1, \beta_2)$ that ranges smoothly over the entire range of unfairness would help in comprehensively profiling quality indicators. Moreover, the unfairness value of an EM-SMPSO algorithm is not absolute in itself \emph{i.e.,} multiple parameter schemes could result in the same value of unfairness. A thorough assessment could enable the creation of selection mechanisms, density estimators, alternate notions of elitism tailored to the usage of EM in swarm-based MOO algorithms.


\end{document}